\documentclass[10pt,twocolumn,letterpaper]{article}

\usepackage{cvpr}              

\usepackage{graphicx}
\usepackage{amsmath}
\usepackage{amssymb}
\usepackage{booktabs}
\usepackage{algorithm}
\usepackage{algorithmic}
\usepackage{mathtools}
\usepackage{dsfont}
\usepackage{multirow}

\usepackage[pagebackref,breaklinks,colorlinks]{hyperref}

\usepackage[capitalize]{cleveref}
\crefname{section}{Sec.}{Secs.}
\Crefname{section}{Section}{Sections}
\Crefname{table}{Table}{Tables}
\crefname{table}{Tab.}{Tabs.}

\def\cvprPaperID{*****} 
\def\confName{CVPR}
\def\confYear{2022}

\begin{document}

\title{Multi-Scale Features and Parallel Transformers Based Image Quality Assessment}

\author{Abhisek Keshari*\\
IIT Jammu\\
{\tt\small 2018ume0126@iitjammu.ac.in}
\and
Komal*\\
IIT Jammu\\
{\tt\small 2020pcs2024@iitjammu.ac.in}
\and 
Sadbhawna\\
IIT Jammu\\
{\tt\small 2018rcs0013@iitjammu.ac.in}
\and 
Badri Subudhi\\
IIT Jammu\\
{\tt\small subudhi.badri@iitjammu.ac.in}
}
\maketitle

\begin{abstract}
With the increase in multimedia content, the type of distortions associated with multimedia is also increasing. This problem of image quality assessment is expanded well in the PIPAL dataset, which is still an open problem to solve for researchers. Although, recently proposed transformers networks have already been used in the literature for image quality assessment. At the same time, we notice that multi-scale feature extraction has proven to be a promising approach for image quality assessment. However, the way transformer networks are used for image quality assessment until now lacks these properties of multi-scale feature extraction. We utilized this fact in our approach and proposed a new architecture by integrating these two promising quality assessment techniques of images. Our experimentation on various datasets, including the PIPAL dataset, demonstrates that the proposed integration technique outperforms existing algorithms. The source code of the proposed algorithm is available online: \href{https://github.com/KomalPal9610/IQA}{https://github.com/KomalPal9610/IQA}.
 
\end{abstract}

\section{Introduction}
\footnote{* indicates that the authors have an equal contribution in the work.}
 In recent years, IQA(Image Quality Assessment) gained a lot of attention because image quality is the key factor for various image-based applications such as Image Restoration(IR), Quality Benchmarking \cite{LIN_iqa_survey,article8}. To calculate the perceptual quality of an image, there is a requirement of an automatic method that can be directly linked with the human perception. 
Full-Reference (FR), No-Reference (NR), and Reduced-Reference (RR) algorithms are the three types of IQA algorithms. In FR, the quality of an image is predicted by comparing the properties or features of target image with its reference image. While in RR and NR algorithms some and no information about the reference image is available. In general, FR algorithms are performing better than the NR images but NR algorithms are preferred in real-time scenario.

Over the years, several IQA metrics have been proposed by different researchers. The most well-known and traditional IQA metrics are mean-squared error (MSE)\cite{MSE}, peak signal-to-noise ratio (PSNR), and SSIM \cite{1284395}. SSIM tries to anticipate the perceptual quality score based upon the structure similarity between the reference and distorted images. \\ A few researchers have used natural scene statistics (NSS) such as MSCN coefficients, image entropy, features based on Benford’s law and energy subband ratio for the purpose of quality assessment\cite{brisque, niqe, 8803047,6172573}. BRISQUE (dubbed blind/referenceless image spatial quality
evaluator)\cite{brisque} IQA method only uses the pixel information of an image to extract the features. BRISQUE uses the normalized luminance coefficients and pairwise products of these coefficients of the spatial natural scene statistics (NSS) model in the spatial domain.

\begin{figure*}[!ht]
\centering
\subfloat[Reference Image]{
  \includegraphics[width=27mm]{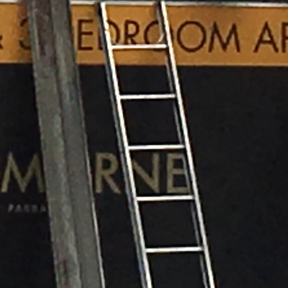} }
  \subfloat[Distorted Image 1 \\ MOS: 0.431]{
  \includegraphics[width=27mm]{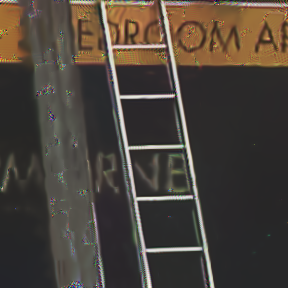} }
  \subfloat[Distorted Image 2 \\ MOS: 0.521]{
  \includegraphics[width=27mm]{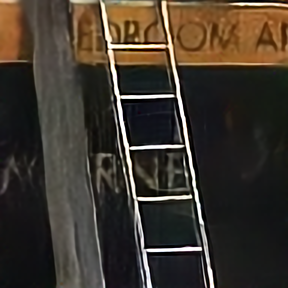} }
  \subfloat[Distorted Image 3 \\ MOS: 0.702]{
  \includegraphics[width=27mm]{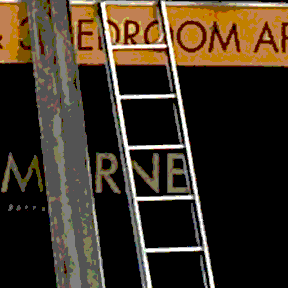} }
  \subfloat[Distorted Image 4 \\ MOS: 0.449]{
  \includegraphics[width=27mm]{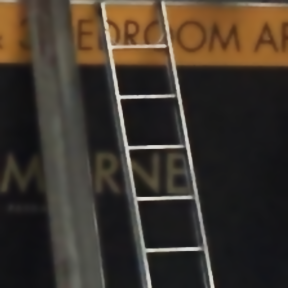} }
  \subfloat[Distorted Image 5 \\ MOS: 0.320]{
  \includegraphics[width=27mm]{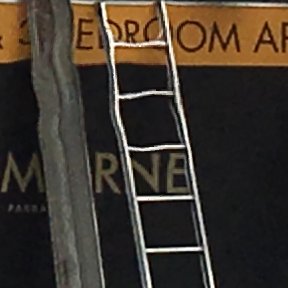} }
\caption{Example reference and distorted Images from PIPAL Dataset. \cite{pipal_dataset}}
\end{figure*}

NIQE \cite{niqe} is also a most popular machine learning-based algorithm for IQA. Without any display of distorted images and any training on distorted images with human opinion scores, NIQE \cite{niqe} mainly uses recognizable deviations from statistical regularities observed in natural images. 
Several researchers have also used gradient information for the purpose of quality assessment such as: \cite{lin_gradient, gmsd}. To calculate the change in contrast and structure of the image, in \cite{lin_gradient} authors have proposed a gradient based method. These gradients are then pooled using component and spatial pooling. Gradient Magnitude Similarity Deviation (GMSD) \cite{gmsd} is based upon predicting the local quality map using the global variation in gradients of reference and distorted images. Further, a global pooling is proposed using this gradient map to calculate the final quality score.

With the development of large datasets (such as TID \cite{TID2013}, KADID \cite{KADID-10k}, PIPAL \cite{pipal_dataset}), CNN-based IQA methods have recently attracted significant attention since convolution neural network(CNN) based state-of-the-art methods are used in many image processing and computer vision applications \cite{9022237}\cite{DBLP:journals/corr/RussakovskyDSKSMHKKBBF14}.
In \cite{9022237}, the quality assessment is done by using Siamese architecture in such a way that cross-dataset performance is not suffered. And by adding
low-level quality cues such as, sharpness, tone and colourfulness, etc.
Sebastian Bosse \MakeLowercase{\textit{et al.}} \cite{FRNR} proposed a CNN-based image quality assessment method that can be used for both FR and NR image quality assessment. This method does not depend on any hand-crafted features or images statistics. An end-to-end neural network with a number of convolutional and pooling layers is used for feature extraction. By cross-database evaluation of this method, the learned features are extremely robust.
Deep similarity for image quality assessment (DeepSim) \cite{DeepSim} measures the local similarities of features of distorted and reference images. To determine the overall quality score, the local quality indices are moderately pooled together.
Deep Image Quality Assessment (DeepQA) \cite{DDD}, is designed without using a human-oriented perspective, and it learns data distribution by various IQA datasets. Feature Pooling Networks(FPN) constructs high-level semantic feature maps at various scales. FPN adds upsampling procedures to the featurized image pyramid \cite{article17} architecture to extract semantically stronger features. Different fields have used these kinds of network designs like Multi-scale multi-view feature aggregation (MSMVFA) \cite{MSMV}. It also combines mid-level attribute features, deep visual features, and high-level semantic features into a single representation for food recognition.


In recent years,  transformer networks \cite{article13} have received a lot of attention due to it's better performance as compared to conventional CNN models. The transformers has also been used in the field of evaluating image quality and shown cutting-edge performance. For example, You \MakeLowercase{\textit{et al.}} proposed the use of vision transformer\cite{article16} for No-Reference IQA \cite{article14}, in which features are extracted using the ResNet-50. Ke \MakeLowercase{\textit{et al.}} \cite{multi_scale_transformers} also used transformer for the purpose of No-reference quality assessment of images. In this paper, authors have used the images at different scales and aspect ratios as the input to the transformer networks and named this network as MUSIQ (Multi-Scale Image Quality Transformer). MUSIQ has the ability to capture the image features at different granularities which makes this network work. In \cite{leida_transformers}, authors have utilised the transformer networks and contrastive loss to catch the features which are quality-aware for the purpose of no-reference quality assessment of images.
The winner of the NTIRE 2021 challenge winners have also used the transformer in Full-Reference IQA \cite{cvpr_ntire_rank-1}. In this paper they have extracted the perceptual features from a CNN backbone. Then these features of reference and distorted images are fed into the encoder and decoder of the transformer for the purpose of evaluating image quality.

In this paper, we proposed a MultiScale transformer-based IQA which is an Full-Reference IQA approach. We named our proposed method as Multi-Scale Features and Parallel Transformers(MSFPT) based quality assessment of
images. MSFPT is specially designed to capture GAN-based distortions which are introduced by PIPAL dataset\cite{pipal_dataset}. Some examples of reference and distorted images in PIPAL dataset are shown through Fig. 1.  Inspired by multi-scale image approaches, we extract the image's features in four different scales by the CNN model. Then these multi-scale features are fed into individual transformers at each scale. The transformer architecture and parameters for all scales are identical. The proposed transformer-based model is then trained for all scales to reliably predict perceptual quality.
\newline
To summarize, the following are our key contributions:
\begin{itemize}
  \item We proposed a new architecture by integrating multi-scale feature extraction and parallel transformers for quality assessment of images.
  \item Our method significantly outperforms previous existing methods on benchmark datasets LIVE \cite{LIVE}, TID2013 \cite{TID2013}, and KADID-10k \cite{KADID-10k}. Also, proposed MSFPT has comparable performance on PIPAL dataset\cite{pipal_dataset} when evaluated as part of NTIRE 2022 IQA Challenge.
\end{itemize}

The remaining paper is organised as: the proposed MSFPT IQA method is elaborated in Section 2, a detailed comparison is conducted on various IQA datasets in Section 3 followed by concluding remarks in Section 4. 

\begin{figure*}[ht]
\centering
\subfloat{
  \includegraphics[width=180mm]{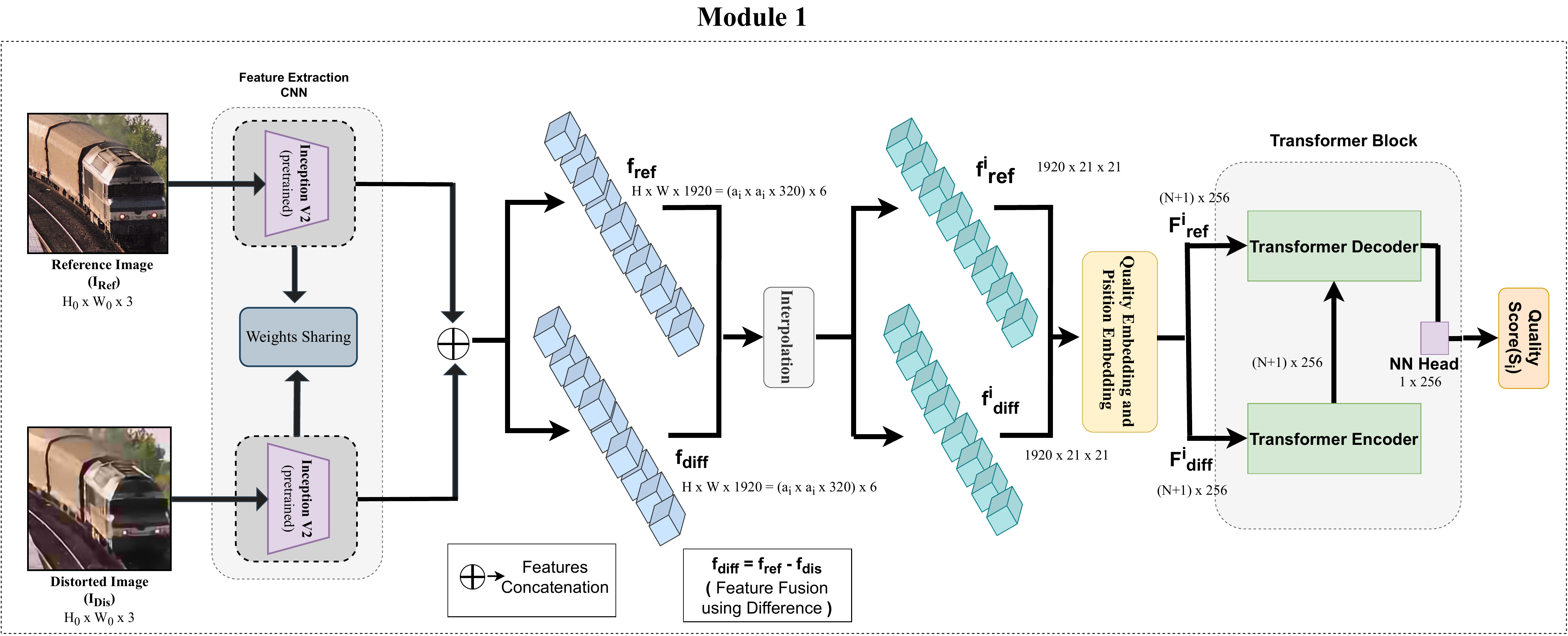} }
\caption{Workflow Diagram of the proposed Module 1.}
\end{figure*}

\begin{figure}[ht]
\centering
\subfloat{
  \includegraphics[width=80mm]{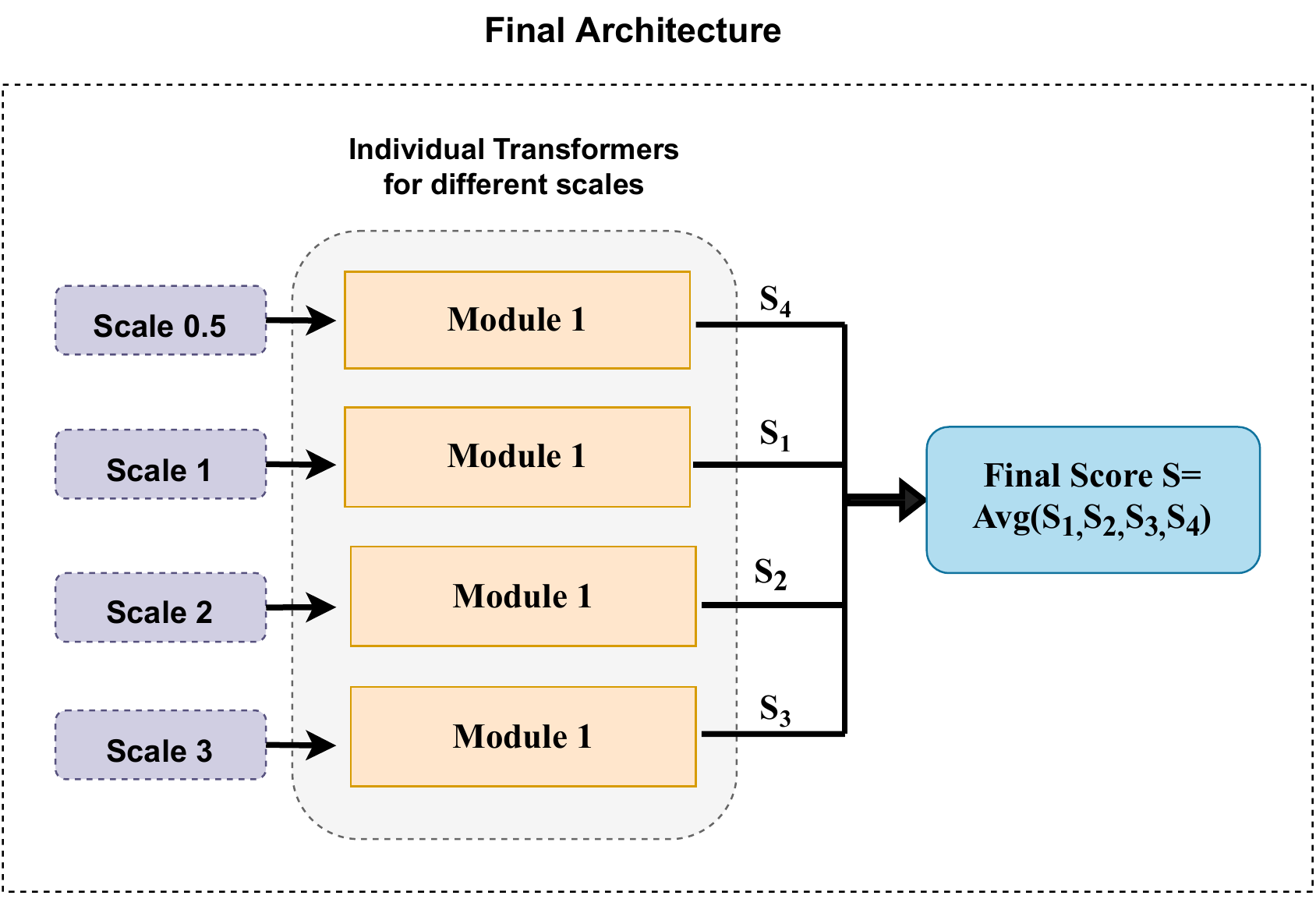} }
\caption{Workflow Diagram of the proposed overall model.}
\end{figure}

\section{Proposed Method}

In this section, we proposed a Multi-Scale Features and Parallel Transformer(MSFPT) network based on NTIRE 2021 challenge winner i.e FR Reference IQA with transformers\cite{cvpr_ntire_rank-1}. The MSFPT network takes pairs of image patches as input. Our proposed method follows multi-scale image quality assesment, via traning four independent model for four different scales of PIPAL dataset images, Scale 1(original image), Scale 2(down-scaled image by factor of 2), Scale 3(down-scaled image by factor of 3) and Scale 0.5(up-scaled image by factor of 2). Multi-scaling is used to analyse the image's GAN-based distortion at different scales. It captures GAN-based texture level noises; hence the multi-scale analysis is critical for image quality assessment \cite{Guo_2021_CVPR}.

Our proposed model consist of four components, Feature extraction block, Interpolation block, Transformer Block, and Averaging Block. 
\textit{Algorithm 1} is the brief psuedo-code of the proposed algorithm. We have also shown the architecture of proposed algorithm through Fig 2 and Fig 3.

\begin{algorithm}
\caption{MultiScale Transformer based IQA}
    
    \hspace*{1em}\textbf{Input:}\text{ A pair of reference R$_{img}$ and distored D$_{img}$ image}
    \newline
        \hspace*{1em}\textbf{Output:}\text{ A predicted IQA score}
    \newline
    \newline
        \hspace*{1em}\textit{Denotes feature extraction as FE,}
    \newline
        \hspace*{1em}\textit{enc\_inp\_emb} =$\{$x$_{ij}$, where i$\in$ $\{$1 ... BatchSize$\}$, j$\in$ $\{$1 ... \hspace*{1em}SequenceLength$\}$, x$_{ij}$=1 $\}$,
        \newline
        \hspace*{1em}\textit{dec\_inp\_emb} =$\{$x$_{ij}$, where i$\in$ $\{$1 ... BatchSize$\}$, j$\in$ $\{$1 ... \hspace*{1em}SequenceLength$\}$, x$_{ij}$=1 $\}$,
    \newline
        \hspace*{1em}\textit{Denotes Transformer block as TB}
    \newline
    \begin{algorithmic}
    \FOR{$j \gets 1$ \textbf{to} $4$} 
    \item
        \hspace*{1em}f$_{ref{_j}}$, f$_{diff{_j}}$ $\coloneqq$ FE(R$_{img}$, D$_{img}$, Scale=j)\newline
        \hspace*{1em}f$_{ref{_j}}^{i} \coloneqq$ Interpolate(f$_{ref{_j}}$)\newline
        \hspace*{1em}f$_{diff{_j}}^{i} \coloneqq$ Interpolate(f$_{diff{_j}}$)\newline
        \hspace*{1em}S$_j \coloneqq$ TB(f$_{ref{_j}}^{i}$, enc\_inp\_emb, f$_{diff{_j}}^{i}$, dec$\_$inp$\_$emb)
    \ENDFOR
    
    Final Score $\coloneqq$ Avg(S$_1$,S$_2$,S$_3$,S$_4$)
    \end{algorithmic}
\end{algorithm}

\subsection{Feature Extraction block}
Similar to \cite{cvpr_ntire_rank-1}, InceptionNet-V2 CNN model \cite{article15}, pre-trained on Image-Net \cite{5206848}, is used as a backbone to extract features. Pre-trained weights are imported and frozen. Intermediate layers, namely \textit{block35\_2, block35\_4, block35\_6, block35\_8, block35\_10, and mixed\_5b} are used as a feature map\cite{Guo_2021_CVPR}.  These blocks are of the same shape for respective scale values, i.e. 
320$\times a_{i} \times a_{i}$, where $a_{i} \in \{33,21,15,9\}$ for scale values i$\in$\{0.5,1,2,3\} respectively. The output of these six feature blocks of the CNN model is concatenated and used as a feature map for the transformer. Pair of   Reference and the distorted image is fed to the backbone model via a bilateral branch\cite{bilateral-branch}. It gives two feature maps as an output, \textit{f$_{ref}$} and \textit{f$_{diff}$}, where \textit{f$_{ref}$} is the feature vector extracted from the reference image and \textit{f$_{diff}$} is acquired from the difference information between reference and distorted images i.e.

\begin{equation}
    f_{diff} = f_{ref} - f_{dist}
\end{equation}

\subsection{Interpolation Block}

 Feature volumes extracted from the above method have a different shape for respective scale values. To process these feature volumes into the transformer, we need a constant shape of 1920$\times$21$\times$21. Using the bilateral interpolation method, we translate the features from different scales (that are 33$\times$33 for Scale 0.5, 15$\times$15 for Scale 2 and 9$\times$9 for Scale 3) to match 21$\times$21.

\subsection{Transformer}

The features extracted from the previous stage are fed into the transformer block. A transformer is a ubiquitous, and recently popular deep learning architecture which works on the principle of self-attention mechanism, weighing the importance of each part of the input data in a deferential manner. The transformers has been successfully used to determine the quality of an image. Many researchers \cite{article14, article13, article16} have reported the use of transformer for image quality assessment. The attention is the fundamental concept that help in improving the performance of neural machine translation applications in a transformer block. Transformers are primarily developed to operate sequential input data. The transformer's attention layer has access to all past states and weighs them according to a learnt measure of relevance, providing relevant information about tokens that are far away. The Transformer employs the architecture with an encoder and decoder network by using the concepts of attention mechanisms and improving parallelization. The output of a transformer is calculated using a weighted average of the values, where the weights for each value are determined by the query's compatibility function with the relevant key. In the proposed scheme we have used the parallel transformers, corresponding to the multi-scale features obtained from at each scale.
\textbf{Transformer encoder},
The difference feature embeddings \textit{F$_{d}$} $\in$ 
\begin{math}
    \mathbb{R}^{N\times D}
\end{math}
, N is number of patches and D is the transformer input dimension, is used as the transformer encoder's input. We begin by reducing the vector \textit{F$_{d}$} dimension to D using 1$\times$1 convoluion layer, followed by flattening of dimensions. The number of patches is determined as \textit{N = W$\times$H}. We append \textit{F$_{d_{0}}$} to the starting of the input feature embedding to add extra quality embedding as used in others vision transformer models \cite{article14,article16}. In order to keep the positional information, the trainable position embedding \textit{P$_{d}$} $\in$ 
\begin{math}
    \mathbb{R}^{(1+N)\times D}
\end{math}
is also incorporated. The encoder's calculation can be expressed as shown below:

\begin{equation}
y_{0} = \{ F_{di}+R_{di}, i\in \{ 0,1,...,N \} \}, \end{equation}
and 
\begin{equation}
q_{i} = k_{i} = v_{i} = y_{i-1},
\end{equation}
and
\begin{equation}
y_{i}^{'} = LN ( MHA ( q_{i},k_{i},v_{i}  ) + y_{i-1} ),
\end{equation}
where 
\begin{equation}
y_{i} = LN(MLP(y_{i}^{'}) + y_{i}^{'}), i \in \{ 1,2,....,L \},
\end{equation} 
and 
\begin{equation}
\left \{ F_{Ei}, i \in \left \{ 1,2,....,N \right \} \right \} = y_{L}, \\
\end{equation}
where $L$ is the number of encoder layers. The input feature embeddings and output has the same size \textit{$F_{e}$} $\in$ 
\begin{math}
    \mathbb{R}^{(1+N)\times D}
\end{math}
.\\

\textbf{Transformer decoder} The decoder takes three components as input the output of encoder \textit{ $F_{E}$}, the reference feature embeddings  \textit{F$_{r}$} $\in$ 
\begin{math}
    \mathbb{R}^{(1+N)\times D}
\end{math},
obtained through reduction followed by flattening, extra quality embeddings and position embedding. \textit{F$_{E}$} is utilised as key-value in second  Multi head attention layer. The calculation of decoder can be formed as:

\begin{table*}[t]
\centering
\renewcommand{\arraystretch}{1.3}
\caption{A tabulated summary of the datasets used for the performance comparison.}\label{tab5: }
\scriptsize
\begin{tabular}{lcccccrrr}\toprule
\textbf{Database} & \textbf{Reference Images} & \textbf{Distorted Images} & \textbf{Distortion Types} & \textbf{Ratings} & \textbf{Rating Type} & \textbf{Distortion Type} & \textbf{Environment} \\\midrule
LIVE\cite{LIVE} &29 &779 &5 &25k &MOS &traditional &lab \\
TID2013\cite{TID2013} &25 &3000 &25 &524k &MOS &traditional &lab \\
KADID-10k\cite{KADID-10k} &81 &10.1k &25 &30.4k &MOS &traditional &crowdsourcing \\
PIPAL\cite{pipal_dataset} &250 &29k &40 &1.13m &MOS &trad. + algo outputs &crowdsourcing \\
\bottomrule
\end{tabular}
\end{table*}

\begin{equation}
y_{0} = \left \{ F_{i} + P_{i}, \forall i \in \left \{ 1,2,....,N \right \} \right \} ,
\end{equation}
and
\begin{equation}
 v_{i} = q_{i} = k_{i} = z_{i-1},
\end{equation}
and
\begin{equation}
y_{i}^{'} = LN\left ( MLA\left ( q_{i},k_{i},v_{i} \right ) + z_{i-1} \right ), \end{equation}
where
\begin{equation}
k_{i}^{'} = v_{i}^{'} = y_{L}, \end{equation},
\begin{equation}
q_{i}^{'} = z_{i}^{'},\end{equation}
\begin{equation}
z_{i}^{"} = LN(MHA(q_{i}^{'},k_{i}^{'},v_{i}^{'}) + z_{i}^{'}),
\end{equation}
and 
\begin{equation}
z_{i}^{'} = LN(MLP(z_{i}^{"}) + z_{i}^{"}), i \in \left \{ 1,2,....,L \right \},
\end{equation}
and 
\begin{equation}
\left \{ F_{Di}, i \in \left \{ 1,2,....,N \right \} \right \} = z_{L},
\end{equation}
where $L$ is the number of encoder layers. The input feature embeddings and output has the same size $F_{E} \in R^{(1+N)\times D}$.

\textbf{Head}. The Neural Network block calculates the final quality score. The NN Block receives the first vector of the decoder output, $F_{D_{0}}$ $\in$ $R^{1\times D}$  in Eq. 2, which carries the quality information. The Block is made up of two completely connected (FC) layers, with the ReLU activation occurring after the first FC layer. A single score is predicted by the second FC layer, which contains one channel.

\subsection{Averaging Module}
Transformer Block $T_i$ predicts the quality score for scale i ($S_i$). The final quality score ($S$) is calculated by averaging the estimated quality score for each scale:

\begin{equation}
Final Quality Score(S) = \frac{\sum_{i=1}^{4}S_{i}}{4}.
\end{equation}

\section{Experiments}

\begin{table}[!t]
\centering
\renewcommand{\arraystretch}{1.3}
\caption{Performence comparison over LIVE\cite{LIVE} and TID2013\cite{TID2013} Datasets.\cite{Table2}}\label{tab6: }
\scriptsize
\begin{tabular}{lrrrrrrr}\toprule
\multirow{0}{0pt}{\textbf{Method}} &\multicolumn{3}{c}{\textbf{LIVE}} &\multicolumn{3}{c}{\textbf{TID2013}} \\\cmidrule{2-7}
&\textbf{PLCC} & \textbf{SRCC} & \textbf{KRCC} & \textbf{PLCC} & \textbf{SRCC} & \textbf{KRCC} \\\midrule
PSNR &0.865 &0.873 &0.68 &0.677 &0.687 &0.496 \\
SSIM\cite{1284395} &0.937 &0.948 &0.796 &0.777 &0.727 &0.545 \\
MS-SSIM\cite{Wang2003MultiscaleSS} &0.94 &0.951 &0.805 &0.83 &0.786 &0.605 \\
VSI\cite{vsi} &0.948 &0.952 &0.806 &0.9 &0.897 &0.718 \\
MAD\cite{mad} &0.968 &0.967 &0.842 &0.827 &0.781 &0.604 \\
VIF\cite{1576816} &0.96 &0.964 &0.828 &0.771 &0.677 &0.518 \\
FSIMc\cite{5705575} &0.961 &0.965 &0.836 &0.877 &0.851 &0.667 \\
NLPD\cite{nlpd} &0.932 &0.937 &0.778 &0.839 &0.8 &0.625 \\
GMSD\cite{gmsd} &0.957 &0.96 &0.827 &0.855 &0.804 &0.634 \\
WaDIQaM\cite{article6} & &0.947 &0.791 &0.834 &0.831 &0.631 \\
PieAPP\cite{PieAPP} &0.908 &0.919 &0.75 &0.859 &0.876 &0.683 \\
LPIPS\cite{article8} &0.934 &0.932 &0.765 &0.749 &0.67 &0.497 \\
DISTS\cite{DISTS} &0.954 &0.954 &0.811 &0.855 &0.83 &0.639 \\
SWD\cite{SWD} &- &- &- &- &0.819 &0.634 \\
IQT\cite{cvpr_ntire_rank-1} &- &0.97 &0.849 &0.943 &0.899 &0.717 \\
IQT-C\cite{cvpr_ntire_rank-1} &- &0.917 &0.737 &- &0.804 &0.607 \\
MSFPT-1 &0.962 &0.976 &0.874 &\textbf{0.955} &\textbf{0.949} &\textbf{0.807} \\
MSFPT-2 &0.958 &0.964 &0.846 &0.872 &0.857 &0.673 \\
MSFPT-3 &0.944 &0.955 &0.824 &0.853 &0.828 &0.635 \\
MSFPT-0.5 &0.963 &0.976 &\textbf{0.875} &0.831 &0.796 &0.598 \\
MSFPT-avg &\textbf{0.972} &\textbf{0.977} &0.874 &0.929 &0.92 &0.752 \\
\bottomrule
\end{tabular}
\end{table}

\begin{table}[t]
\centering
\renewcommand{\arraystretch}{1.3}
\scriptsize
\caption{Performence comparison over KADID Dataset.\cite{KADID-10k}}\label{tab7: }
\begin{tabular}{lrrr}
\hline
\multirow{-1}{0pt}{ \textbf{Method}} &\multicolumn{3}{c}{ \textbf{KADID}} \\ \cmidrule{2-4}
&\textbf{PLCC} & \textbf{SRCC} & \textbf{KRCC} \\ \midrule
SSIM\cite{1284395}            & 0.723                              & 0.724                               & 0.537                               \\ 
MS-SSIM\cite{Wang2003MultiscaleSS}          & 0.801                              & 0.802                               & 0.609                               \\ 
IWSSIM\cite{Iwssim}          & 0.846                              & 0.850                                & 0.666                               \\
MDSI\cite{mdsi}            & 0.873                              & 0.872                               & 0.682                               \\ 
VSI\cite{vsi}             & 0.878                              & 0.879                               & 0.691                               \\ 
FSIM\cite{5705575}            & 0.851                              & 0.854                               & 0.665                               \\ 
GMSD\cite{gmsd}            & 0.847                              & 0.847                               & 0.664                                 \\
SFF\cite{SFF}             & 0.862                              & 0.862                               & 0.675                               \\ 
SCQI\cite{SCQI}            & 0.853                              & 0.854                               & 0.662                               \\ 
ADD-GSIM\cite{ADD-GSIM}        & 0.817                              & 0.818                               & 0.621                               \\
SR-SIM\cite{SR-SIM}          & 0.834                              & 0.839                               & 0.652                               \\
MSFPT-1     & 0.822                              & 0.846                               & 0.653                               \\ 
MSFPT-2     & 0.796                              & 0.799                               & 0.613                               \\ 
MSFPT-3     & 0.667                              & 0.674                               & 0.495                               \\
MSFPT-0.5   & 0.857                              & 0.857                               & 0.672                               \\
MSFPT-avg   & \textbf{0.888}                     & \textbf{0.883}                      & \textbf{0.700}                        \\ \hline
\end{tabular}
\end{table}


\begin{table}[t]
\renewcommand{\arraystretch}{1.3}
\centering
\label{tab1: }
\scriptsize
\caption{Ablation study with respect to the different scales.}
\begin{tabular}{lrrrr}\toprule
\multirow{-0.5}{0pt}{\textbf{Model Name}} &\multicolumn{3}{c}{\textbf{Validation}} \\\cmidrule{2-4}
& \textbf{Main Score} & \textbf{PLCC} & \textbf{SRCC} \\\midrule
MSFPT-1 &1.552 &0.784 &0.768 \\
MSFPT-2 &1.522 &0.773 &0.749 \\
MSFPT-3 &1.47 &0.749 &0.721 \\
MSFPT-0.5 &- &- &- \\
MSFPT-avg &1.598 &0.810 &0.788 \\
\hline
\multirow{-0.5}{0pt}{\textbf{Model Name}} & \multicolumn{3}{c}{\textbf{Testing}} \\ \cmidrule{2-4}
& \textbf{Main Score} & \textbf{PLCC} & \textbf{SRCC} \\\midrule
MSFPT-avg &1.450 &0.738 &0.713 \\
MSFPT-1 &1.254 &0.637 &0.617 \\
MSFPT + Bert + Scale1 &1.383 &0.699 &0.684 \\
MSFPT + Bert + Scale2 &1.361 &0.698 &0.663 \\
MSFPT + Bert + Scale3 &1.182 &0.621 &0.561 \\
MSFPT + Bert + Avg. of 1,2,3 &1.44 &0.73 &0.71 \\
\bottomrule
\end{tabular}
\end{table}

\begin{table}[t]
\renewcommand{\arraystretch}{1.3}
\centering
\scriptsize
\label{tab2: }
\caption{Performance comparison of the proposed algorithm in NTIRE IQA challenge, Testing phase. }
\begin{tabular}{cccc}
\hline
\textbf{Team Name}  & \textbf{Main score} & \textbf{PLCC}  & \textbf{SRCC}  \\
\hline
Anynomus1      & 1.651      & 0.826 & 0.822 \\
Anynomus2      & 1.642      & 0.827 & 0.815 \\ 
Anynomus3      & 1.64       & 0.823 & 0.817 \\ 
Anynomus4      & 1.541      & 0.775 & 0.766 \\
Anynomus5      & 1.538      & 0.772 & 0.765 \\ 
Anynomus6      & 1.501      & 0.763 & 0.737 \\
\textbf{Pico Zen(ours)} & 1.45       & 0.738 & 0.713 \\ 
Anynomus8      & 1.403      & 0.703 & 0.701 \\ 
\bottomrule
\end{tabular}
\end{table}

\begin{figure*}[t]
\centering
\subfloat[LIVE Dataset]{
  \includegraphics[width=70mm,height=43mm]{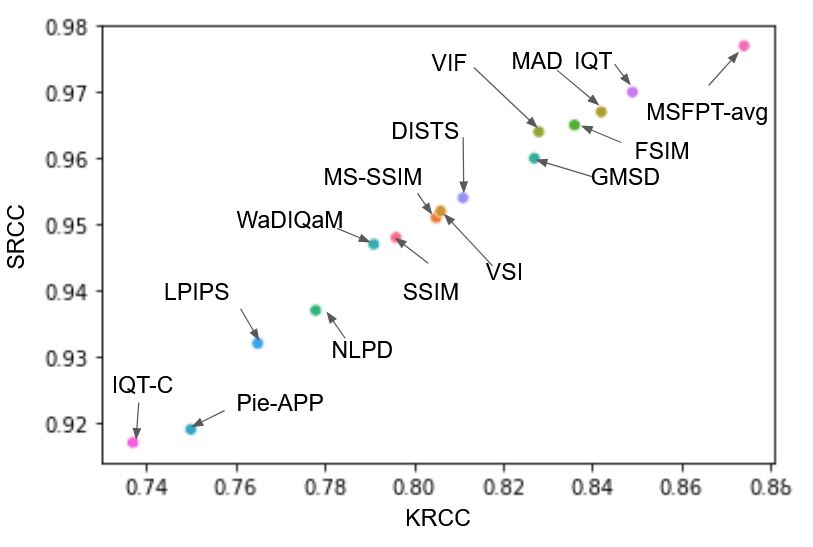} }
  \subfloat[TID 2013 Dataset]{
    \includegraphics[width=70mm]{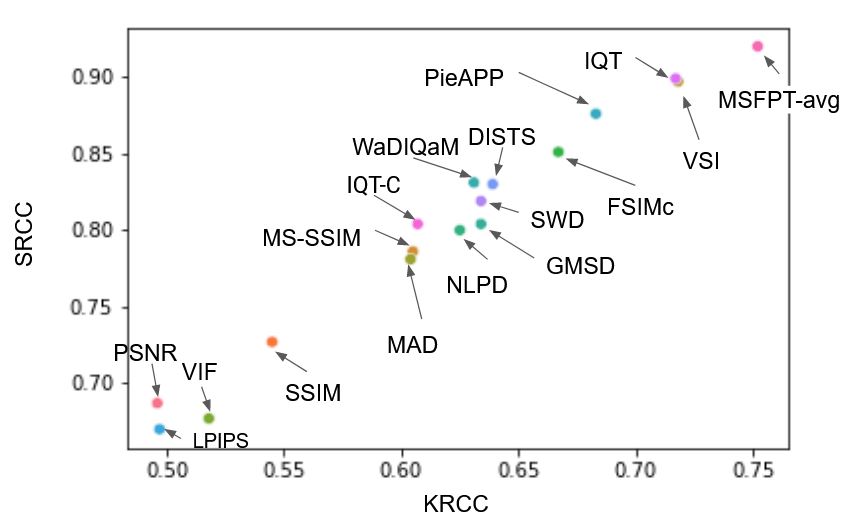} }
    \hspace{0mm}
    \subfloat[KADID Dataset]{
  \includegraphics[width=70mm,height=43mm]{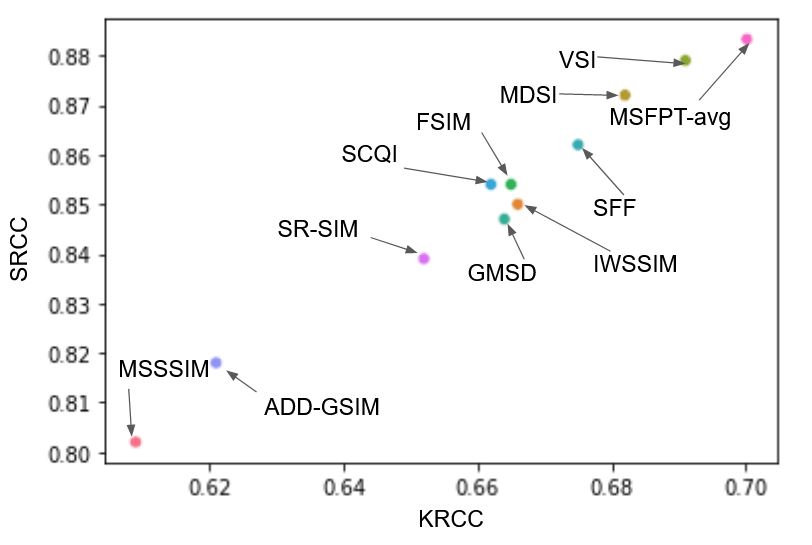} }
  \subfloat[PIPAL Dataset]{
  \includegraphics[width=70mm,height=44mm]{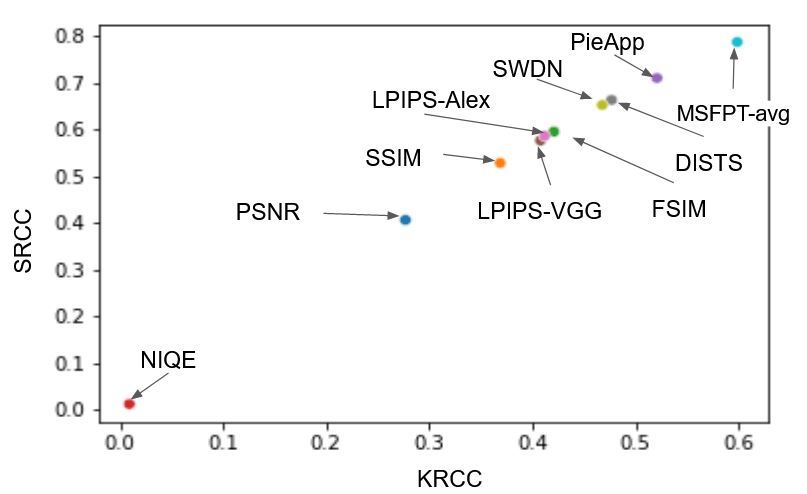} }
\caption{Quantitative comparison of IQA methods. (a) LIVE Dataset, (b)TID 2013 Dataset, (c)KADID Dataset, (d) PIPAL Dataset.}
\end{figure*}

\subsection{Datasets}
Our experiments are conducted on four benchmark Image quality datasets, LIVE \cite{LIVE}, TID2013 \cite{TID2013}, KADID-10k \cite{KADID-10k} and PIPAL \cite{pipal_dataset}. The LIVE dataset contains 29 reference images; from these images, using five different traditional distortion types, 779 distorted images are created. TID2013 contains 25 reference images as well as 3,000 distorted images generated by 24 different distortions, with five levels for each distortion type. KADID-10k includes 81 reference images and 10.1k distorted images generated by 25 distortions. PIPAL contains 250 reference images and 29k distorted images generated by 40 kinds of distortions. This dataset has traditional and algorithmic outputs, i.e. GAN-based distortions produced by different GAN based algorithms.
The validation set of the PIPAl dataset contains 25 reference images and 1650 distortion images, whereas the testing set of the PIPAl dataset contains 25 reference images and 1,650 distortion images.

\subsection{Implementation Details}
In the training phase, a given image is cropped to obtain the desired patch of size HxWxC. For PIPAL dataset we have H = W = 192, C = 3 and for LIVE\cite{LIVE}, KADID-10k\cite{KADID-10k} and TID2013\cite{TID2013} H = W = 256, C = 3.
The feature volume of MSFPT has N = 442 patches. In testing phase, same number of patches are obtained from the image pair given. We extract M overlapping patches where M is the number of ensembles used and use an average of M individual patch quality ratings to predict the final quality score.
The Adam optimizer was used with weight-decay $\alpha$ = 1$e^{-5}$, $\beta_{1}$ = 0.9 and $\beta_{2}$ = 0.999 with L1 as a loss function since it is more resilient to outliers than MSE loss. We have set the learning rate to 2$e^{-4}$ and used cosine annealing learning rate scheduler,  A batch size of 16 was chosen. PyTorch 1.10.1 was used with two NVIDIA V100 GPUs and CUDA 11.0. 
Data augmentation, including random crop, vertical flip, random rotation, and horizontal flip, is applied during the training.

We compare MSFPT network with several state-of-the-art methods on all four datasets \cite{LIVE,KADID-10k,TID2013,pipal_dataset} for IQA. The methods have deep learning-based methods such as PieAPP \cite{Prashnani_2018_CVPR}, LPIPS \cite{article8}, SWD \cite{eessIV} and  DISTS \cite{DISTS} and shallow methods like SSIM \cite{1284395} and PSNR.
For most cases our method shows more promising results than current deep learning-based methods. Our model out performs other deep-learning based models like IQT method\cite{cvpr_ntire_rank-1} on LIVE\cite{LIVE} data set by 0.07 SRCC and 0.029 in KRCC. In case of TID2013\cite{TID2013} by using weight sharing and multi-scale we outperform existing deep-learning models by 0.021 SRCC, and 0.034 KRCC. For KADID-10k\cite{KADID-10k}, it outperforms various IQA methods like VSI by 0.01 PLCC, 0.004 SRCC and 0.009 KRCC. 

\subsection{Ablation study}
The use of different information between various scales of input images is one of the vital characteristics in the proposed architecture. Four types of scales are available, i.e. 1, 2, 3 and 0.5, as mentioned in the Table 4. We conducted an ablation experiment to study the influence of input shape and transformer type, and the results of the performance evaluation are provided in Table 4.

In the proposed algorithm, we have used Attention is all you need \cite{article13} transformer, that gives significantly better performance over CNN based models. We have also tried to used Bert in the proposed algorithm and observed (from Table 4) that the Bert is giving slightly poorer performance.  These results clearly validates that incorporating multi-scale features with the parallel transformers significantly improves the performances.

\subsection{ NTIRE 22 IQA Challenge Report}
In both validation and testing phases, we use MSFPT model trained on PIPAL dataset on four different scales with batch size 16. Table 5 shows the competition's final ranking during the testing phase.

\section{Conclusions}
In this paper, we presented a full-reference image quality assessment algorithm based on parallel transformers and multi-scale CNN features. These features are trained for the purpose of quality prediction using transformers network with  encoders and decoders. We conducted extensive experimental studies to show the superiority of using this combination of parallel transformers and multi-scale features as compared to other combination of networks. The proposed method outperforms current state-of-the-art image quality assessment methods in terms of performance.

\begin{table}[!htp]\centering
\caption{Performance comparison over Validation Dataset of NTIRE-2022 FR\cite{gu2022ntire} }\label{tab3: }
\scriptsize
\begin{tabular}{lrrrr}\toprule
\textbf{Model Name} &\textbf{Main Score} &\textbf{SRCC} &\textbf{PLCC} \\\midrule
MSFPT-avg (our) &1.598 &0.81 &0.788 \\
PSNR &0.503 &0.234 &0.269 \\
NQM\cite{nqm} &0.666 &0.302 &0.364 \\
UQI\cite{uqi} &0.966 &0.461 &0.505 \\
SSIM\cite{1284395} &0.696 &0.319 &0.377 \\
MS-SSIM\cite{Wang2003MultiscaleSS} &0.457 &0.338 &0.119 \\
RFSIM\cite{rfsim} &0.539 &0.254 &0.285 \\
GSM\cite{gsm} &0.829 &0.379 &0.45 \\
SRSIM\cite{SR-SIM} &1.155 &0.529 &0.626 \\
FSIM\cite{5705575} &1.005 &0.452 &0.553 \\
VSI\cite{vsi} &0.905 &0.411 &0.493 \\
NIQE\cite{niqe} &0.141 &0.012 &0.129 \\
MA\cite{ma} &0.196 &0.099 &0.097 \\
PI\cite{Pi} &0.198 &0.064 &0.134 \\
Brisque\cite{brisque} &0.06 &0.008 &0.052 \\
LPIPS-Alex\cite{article8} &1.175 &0.569 &0.606 \\
LPIPS-VGG\cite{article8} &1.162 &0.551 &0.611 \\
DISTS\cite{DISTS} &1.243 &0.608 &0.634 \\
\bottomrule
\end{tabular}
\end{table}

\begin{table}[!htp]\centering
\caption{Performance comparison over Testing Dataset of NTIRE-2022 FR\cite{gu2022ntire} }\label{tab4: }
\scriptsize
\begin{tabular}{lrrrr}\toprule
\textbf{Model Name} &\textbf{Main Score} &\textbf{SRCC} &\textbf{PLCC} \\\midrule
MSFPT-avg (our) &1.45 &0.738 &0.713 \\
PSNR &0.526 &0.249 &0.277 \\
NQM\cite{nqm} &0.76 &0.364 &0.395 \\
UQI\cite{uqi} &0.87 &0.42 &0.45 \\
SSIM\cite{1284395} &0.753 &0.361 &0.391 \\
MS-SSIM\cite{Wang2003MultiscaleSS} &0.532 &0.369 &0.163 \\
RFSIM\cite{rfsim} &0.632 &0.304 &0.328 \\
GSM\cite{gsm} &0.874 &0.409 &0.465 \\
SRSIM\cite{SR-SIM} &1.209 &0.573 &0.636 \\
FSIM\cite{5705575} &1.075 &0.504 &0.571 \\
VSI\cite{vsi} &0.975 &0.458 &0.517 \\
NIQE\cite{niqe} &0.166 &0.034 &0.132 \\
MA\cite{ma} &0.287 &0.14 &0.147 \\
PI\cite{Pi} &0.249 &0.104 &0.145 \\
Brisque\cite{brisque} &0.14 &0.071 &0.069 \\
LPIPS-Alex\cite{article8} &1.137 &0.566 &0.571 \\
LPIPS-VGG\cite{article8} &1.228 &0.595 &0.633 \\
DISTS\cite{DISTS} &1.342 &0.655 &0.687 \\
\bottomrule
\end{tabular}
\end{table}

{\small
\bibliographystyle{ieee_fullname}
\bibliography{egbib}
}

\end{document}